\pdfoutput=1

\documentclass[11pt]{article}

\usepackage[]{ACL2023}

\usepackage{times}
\usepackage{latexsym}
\usepackage{linguex}
\usepackage{graphicx}
\usepackage{csquotes}

\usepackage[T1]{fontenc}

\usepackage[utf8]{inputenc}

\usepackage{microtype}

\usepackage{inconsolata}

\newcommand{\fol}{\textsc{fol}}
\newcommand{\lm}{\textsc{lm}}
\newcommand{\nlp}{\textsc{nlp}}

\title{A Primer on Computational Semantics for Artificial Intelligence Systems}

\author{Casey Kennington\\
  Computer Science \\
  Boise State University \\
  \texttt{caseykennington@boisestate.edu} 
  }

\begin{document}
\maketitle

\begin{abstract}
As people adopt transformer-based language models (e.g., ChatGPT and Gemini) for an increasing number of use-cases, it is important to know how such models learn and represent the meaning of the language, and to be more informed about what language is. This document is an attempt to help the reader understand how linguistic meaning (i.e., semantics) is approached from different fields of scientific and philosophical examination. I also explain three primary semantic theories: formal semantics, grounded semantics, and distributional semantics then compare how transformer-based language models differ from how humans learn language. 
\end{abstract}

\section{Introduction}

I often ask my students \enquote{What is the meaning of the word \emph{red}?} They are usually upper-division undergraduate or graduate students who have shown their talents and capabilities in prior coursework, but this question always stumps them. One student usually points out that \emph{red} is a color, but I explain that \emph{color} is a category of the word \emph{red}, not the meaning of \emph{red}. Someone else usually points to something more technical like hexadecimal values, light wavelengths, or rods and cones in the eyes, but when a child learns the word \emph{red}, they don't require complex knowledge to identify red things--kids just do it. Of course, each student with functional eyesight knows the meaning of the word \emph{red}. The challenge is that they can't really explain \textit{how} they know it. \emph{Red} doesn't have a definition because \emph{red} is a \emph{concrete} category: the word \emph{red} refers to things in the world, and we come to a knowledge of the meaning of \emph{red} through experience with things that are denoted as red.\footnote{And definitions aren't meanings. Definitions are attempts at describing connotations.}

The goal of posing questions about colors to my students is to point out that we as humans (at least most of us) take our ability to produce and understand language for granted. If we cannot even explain the meaning of a term so clearly obvious as the word \emph{red}, then what does the average person actually know about the language they speak? Certainly, most people don't understand how car engines work, but that doesn't keep them from driving safely. Analogously, people engage in conversations effortlessly, but don't really have to understand the mechanics of how it all works in order to be effective communicators. That's part of the wonder of language. Now that artificial intelligence models can very effectively use language as a medium to interact with people, what does that mean for language as we understand it? Understanding how language works might help us contend with these models (e.g., when do they break down? should we trust them? should we take their outputs seriously?). We cannot just assume that we know what things mean because we cannot assume that the models capture the meaning in the same way as we was humans understand words, which has implications for what we should and shouldn't ask of these models. 

My focus in this document is on \emph{meaning}, because ultimately we need to understand what meaning is if we are to convey meaning to machines. A broader goal of this paper is to discuss what is known about language from several perspectives, why language matters for people, and, more importantly, what language really means for artificial intelligent agents and how we interact with them. 

\paragraph{Document Plan} Up next, I define language. Then I outline different ways to look at language, from linguistics, psychology, cognitive science, neuroscience, social science, and child development.  I then lay out some of the philosophy of meaning (albeit far from exhaustively). Following that, I explain three different approaches of study that focus on meaning that could be computed by machines: formal, grounded, and distributional approaches. That is followed by a comparison of how humans learn language, human cognition, the role of emotion, then we look at meaning in the current snapshot of AI, mostly focusing on language models and the transformer architecture. 

\section{What is Natural Language?}

Broadly defined, natural language is a means of symbolic communication that occurs naturally in a human community by process of use, repetition and change \cite{Wikipedia-contributors2024-we}. Though not necessarily a requirement (but is nonetheless most often the case), a language evolves over time without conscious planning or predetermination. For example, no committee of English language experts got together to decide to bring the new word \emph{sus} to the community, what it should mean, and that people should start using it in stead of other, perfectly good words that have similar meanings (i.e., \emph{suspicious}). Quite the contrary: kids just started saying \emph{sus} and it caught on in a bottom-up, crowd-driven fashion. New words are added to the English--and many other languages of thew world--lexicon quite routinely, grammatical rules change, and the meanings of words shift as they are used over time. 

But what do I mean by \emph{symbolic} communication? Simply, that natural languages use symbols to convey information. A word is a symbol. If I utter \emph{dog} to someone with the intent of explaining about a dog I had seen earlier, I do not need the listener to actually see the dog that I saw; I can just use the word to symbolize \emph{dog} for the purpose of bringing what I saw to the mind of the listener. Thus, in English at least, the word \emph{dog} whether it is written, spoken, or signed, is a symbol for the category of things in the world that we call dogs. I will use the word \emph{symbol} often below, so what I mean should become clearer. 


Anyone who speaks multiple languages can attest to the messiness of language because languages do not translate cleanly. The French word \emph{si}, is a fun example. A simple, monosylabic word, \emph{si} generally means \emph{yes} in English, but not exactly. It also marks polarity. Let me explain: suppose I ask a colleague if they are coming to a meeting. I can ask in multiple ways: \emph{are you coming to the meeting?} or \emph{are you not coming to the meeting?}. Both convey the same meaning, but in the former question the polarity of the sentence is positive, whereas in the latter, the polarity is negative. In French, if I ask someone something like \emph{are you not coming to the meeting?}, and they answer \emph{si}, they are saying that they in fact will come to the meeting, despite me assuming they wouldn't in a negative way. Doing this in English is more complicated: if I say \emph{yes}, then my answer is still ambiguous. Does \emph{yes} mean that I am agreeing that I will in fact not come to the meeting, or does \emph{yes} mean that I will in fact come to the meeting? We often add information when we respond to negative polarity questions like \emph{yes, I am coming} or \emph{yes, I have a deadline}. In French, one word does the trick and it took an entire paragraph to explain that one, useful word. This example illustrates that the symbols we use to connote concepts can be very different from language to language.

\subsection{Written, Spoken, and Signed Language}

When we think of words in a language, we often think of the written, textual form of the words because that might be the easiest way to convey and recall them. But, text is not language. Text is a \textit{part} of the way we communicate, but many languages don't even have writing systems. Speech or signed symbols come first both in evolution of language and in children learning their first language. I focus a lot more on speech than signed language here, but both are valid methods for conveying information; the former using sound signals across time. For example, when I utter the word \emph{computer} I push air from my lungs through my dynamically changing mouth to speak the word \emph{computer} from the first consonant to the last---which takes time, albeit less than a second. The same goes for signed language, only instead of using vocal vibrations and air through the mouth, information is conveyed visually using hands, arms, etc., as they move over time. 

Language isn't just a thing. Rather, like many objects of interest, it is a \textit{process}. The way we use language is a process, the way meanings change is a process, and the way we pronounce things also is a process. No language is a static thing as long as people are still using it.

In the following subsections, I explain language as it is viewed from a handful of different fields. The explanations obviously do not do justice to the breadth or depth to which research in those fields have given us knowledge about language, but they should at least give us some understanding about different lenses we can use to investigate what language is and how we as people make use of it.

\subsection{Linguistics}

Linguistics is the study of language. I purposefully write ``language" instead of ``languages" because in linguistics any individual language or set of languages is fair game. Linguistics is a field that looks at phenomena underlying language whether spoken, signed, written, or otherwise, as the primary object of intrinsic value. Importantly, linguists tend to be \textit{descriptive} in that they are interested in what people actually do when using language, not \textit{prescriptive} in that they don't want to tell people how they should use language.

Most often, linguistics is broken down into sub-areas where more specific interest can be focused:

\begin{itemize}
    \item Phonetics and phonology - the anatomy of speech production and the sound of language as it is spoken
    \item Morphology - how words morph into different forms depending on how they are used; for example, the word \emph{play} can morph into \emph{plays} or \emph{playing} or \emph{played} depending on what a speaker intends
    \item Syntax - the grammar or structure of language. For example, when we utter sentences, there is a subject and potentially an object. Syntax dictates how subjects are marked (with some kind of determiner or particle, or sentential order); where verbs and objects go. 
    \item Semantics - the meaning of language including lexical meaning (lexical=word or sub-word level meanings, for example the meaning of the word \emph{play} and the sub-word \textit{ed} in English means past-tense, so \textit{played} means to play in the past), as well as meanings of phrases, sentences, or broader uses of language
    \item Pragmatics - the contexts in which language is used and how language means things that are outside of language itself (for example, if someone is pounding nails into the top of an expensive coffee table and you yell \emph{What are you doing?!?}, what you are asking isn't just the words that you yelled, but also \emph{why} they are doing it, and by yelling you are conveying that the action is probably unexpected, and undesired)\footnote{Thanks to Bill Watterson for this example.}
\end{itemize}

These sub-areas of linguistics aren't rigid categories. Linguists interested in morphology might simultaneously touch areas of syntax and lexical semantics, and linguists interested in semantics might find themselves worrying a lot about the role of pragmatics. Moreover, there are other aspects of language that could concern a particular linguist, for example dialogue (how two or more people communicate using speech) which involves all the areas of language listed above, discourse (how language is used beyond simple words and sentences), language documentation (taking steps to make sure examples and characteristics of a language are not lost; for example if a language is endangered), how people use language on social media vs. in person, how new words are introduced to a language speaking community, among other possible objects of study where language is concerned. 

While all areas of language are worth more exploration, my primary focus here is semantics---where meaning is concerned. Why semantics? Because meaning of words, in my view, is very very important for how we communicate with each other. Because somehow a model running on a computer needs to grasp some degree of meaning in order to arrive at meaningful outputs or behavior and use words appropriately. How humans acquire, represent, and use words in a meaningful way is a question for areas like cognitive science, child development, psychology, psycholinguistics, and neuroscience, which I touch on below. How computational models come to acquire, represent and use words in a meaningful way is another thing entirely, discussed throughout this document.

\subsection{Semiotics}

Beyond using words, people send signals to each other in other ways such as head-nodding, body posture, hand signals like waving, and other communicative acts. What makes these signals meaningful to people (or even non-human organisms)? This question is central to the field of \emph{semiotics}. Where linguistics is interested in language, semiotics is interested in any communicative signal \cite{Atkin2023-wd}. 

Within semiotics Charles Pierce distinguishes between the \emph{interpretant} and the \emph{interpreter}. The former is the  mental representation of an object, and the latter is the person who has the interpretant in their head. While the study of linguistics can focus on individual use of language, semiotics treats people who understand a sign as first class citizens. 

Like semantics (and to some degree, pragmatics), semiotics is interested in meaning. In fact, what makes an organism use a sign (e.g., a word, a hand gesture, a smoke signal, or a specific dance), derive and share meaningful signs with others is of prime interest. 

\subsection{Philology \& Etymology}

Philology is focused on language from historical sources. Philology has some overlap with linguistics because a philologist might be interested, for example, in how the syntax of a language split into two over time. However, Philology is also part of History as a discipline, literary criticism, and is interested in Etymology which is the study of the study of words' origins and evolution (including their phonology and semantics) across time. 

When old texts are uncovered in an archaeological site, philologists are often the ones who are called upon to determine which language the text encodes, which period of time it came from, and of course what the text itself means. 

Sometimes linguists employ philology and sometimes philologists employ theories of linguistics. For example a linguist interested in comparing the syntax of two Indo-European languages at a certain point in time might collaborate with a philologist. 

\subsection{Psychology \& Pyscholinguistics} 

Whereas linguistics looks at language as a field of study on its own, psychology and psycholinguistics look at the organisms that use language, as well as what behaviors are related to language. 

Psychology is the study of mind and behavior. Psychology examines things that affect human existence, including human perception, cognition, attention, emotion, intelligence, personality, and relationships. Language is interrelated with many of these things. Clearly, the fact that humans learn and use language has something to do with individual human experience and psychology has a lot to say about individual human experience. 

Psycholinguistics focuses on language within the scientific framework of psychology. Psycholinguistics is also interested in things that affect human existence like perception and attention, but with focus on how they affect language learning, processing, and production. For example, eye tracking studies have looked at how humans look at words as they are read (spoiler alert: we don't just read words in sequential order, our eye gaze jumps around a lot \cite{Blythe2011-fc}), and studies looking at brain activations (e.g. using EEG) have shown what happens when someone encounters a grammatical vs. some kind of pronoun reference surprise when they are reading passages. 

\subsection{Cognitive Science}

Cognitive science is also interested in the study of the human mind and brain, but it focuses more on how the brain manipulates knowledge, and how mental representations and processes happen within the brain itself. While there is overlap between cognitive science and psychology---and both are very interdisciplinary---the outcomes are often different. For example, a psychologist might examine personalities and how they affect human behavior, whereas cognitive science might be more focused on the mental aspects of a specific personality type. 

If that doesn't help distinguish cognitive science from psychology, take heart because many people find it difficult to distinguish between the two fields. The way I explain cognition is \emph{anything you can think about thinking about}. For example, when you see a red object (a shirt, say), the light comes into your eyes at a certain wavelength, which then travels through your retinas, then is picked up by specific cells in your eye for distinguishing that specific wavelength, which then activates parts of your brain, which then primes thoughts of the word \textit{red}, and concepts that relate to \textit{red}. You cannot attend to or think about what is happening at the basic level of perception (i.e., what's happening to your eyes, rods, cones, or individual neurons), but at the point where you can think about what you are perceiving, that is cognition. Everything before that is perception. That's not to say that cognitive scientists aren't interested in how perception affects cognition (quite the contrary!), but the main object of examination is the higher-order thinking that happens within human brains. 

Language, of course (and understanding meaning of words), is something that requires cognition. We can read or listen to other people's words, we can think about those words, and we can respond in our own way to those words, either through a verbal response or some kind of action. We can think about words, and many people often think in words. Language is an important cognitive enterprise, though cognition need not be linguistic in nature. 

\subsection{Neuroscience}

As complex as galaxies and black holes are, one of the most complex things that we as humans have ever encountered sits between our ears: the human brain. Neuroscience is the study of the brain, but also nervous system and spinal cord. Neuroscience is more interested in the anatomy of the brain, how individual neurons work, what different neuron types do, as well as neurological functions including learning, memory, and other things in the realm of psychology and cognitive science, though neuroscience is more generally biological than the other two fields. 

Without the brain, we wouldn't be able to learn language. To use language, we need to hear others speak (or see others sign), memory to hold those experiences, the ability to link symbols to real-world language usages, the ability to control the voice and mouth to generate speech, pick the right words when speaking, learn and apply grammar rules, etc. There have been efforts to determine if there are specific areas of the brain that are more language-related such as Broca's or Wernicke's areas of the brain with varying degrees of success. 

Neuroscience is a fascinating field, and what has been learned about the neuroscience of language has been exciting in recent years with new ways to observe brain functions, such as MRI machines. It's  not quite clear how meaning is stored and retrieved in the brain, and what is meant by meaning? Some have explored this (e.g., \citet{Baggio2018-lp,Pulvermuller1999-rj,Malt2019-ly,Dreyer2018-fc}) but there is a lot of work needed. 

\subsection{Social Science}

Without other people, why would we need language? Even though language must be learned by each individual, one of the primary uses of language is to communicate with other people. This means that language is not just a psychological, cognitive, or neurological system, but also very social. 

The social sciences include various fields such as culture, anthropology, religion, sociology, political science, communication, among others. Each sub-field relates to language in its own way, for example, how culture influences language use, and the other way around; or how ancient humans communicated with each other. 

Social sciences and the concern for language can range from how dialogue takes place between two friends, or between two strangers to how language is used to manipulate a large group. How has the Internet and social media changed the way people communicate with each other? The list of interesting questions goes on. We need social science to go beyond the individual, because that's the setting where language is used. 

Social science has a lot to say about meaning, because language speakers derived and update meanings of words as they interact with other people in social settings. Some argue that meaning is not in our heads, but shared across social groups. There's some truth to that. 

\subsection{Education}

Education is the transmission of knowledge, skills, and character traits and manifests in various forms \cite{Wikipedia-contributors2024-xx}. Education is so important, that many countries dedicate large amounts of public funding to educate their citizens from children to university-level. Language, particularly reading and writing, of course plays a role in education: most learning is done in a setting where language is a medium for transmitting knowledge. But before someone can use written language to learn, they need to first learn how to read and write. One of the first things that is taught to children is \emph{literacy}---the ability to read and write---so children gain the ability to use language to acquire knowledge: "learn to read, read to learn."

Reading and writing are important uses of language for the sake of education itself, but also because reading and writing are basic and critical skills that people need to function in society. Much effort of educational research is put into finding better ways to help children (and adults) learn how to read and write. It is assumed that children can already speak and listen, therefore children understand words before they are able to be taught to read and write, then they are taught the written symbol system of their language.

Children first learn how to pronounce words that they read. Being able to read the words so they are pronounced correctly is known as \emph{decoding}, but another important step goes beyond just knowing how written words are pronounced: \emph{comprehension}. Comprehension means understanding what one is reading. Without comprehension, there really is no point to reading because the meanings of the words are not being transmitted. Comprehension of course means that children also grasp the meanings of the words that they are reading. 

In short, Education is crucial if children are to become literate users of a language's written symbol system which enables those children to become more powerful language users.

With this background, we now turn our attention to meaning of language beginning with philosophy.

\section{Philosophy of Meaning\footnote{Some of the material for this section is taken from Chapter 2 of \citet{Kennington2016-pl}.}}

In this section, I consider linguistic meaning from a philosophical stance. 

\subsection{Sense \& Reference, Connotation \& Denotation, \& Intension and Extension}

To begin understanding the philosophy of meaning, we start with \citet{Abbott2010-gs} and focus on \emph{reference}. By reference, I mean that humans use words to refer to objects, events, or people in the world, or within language itself. Phrases like \emph{the man on the left}, \emph{the sun}, \emph{that thing}, and \emph{it} are all different kinds of \emph{referring expressions}. 

Why start with referring expressions when talking about the philosophy of language? Because (1) referring to physical objects makes up a high proportion of the expressions spoken by children learning their first language and (2) philosophers have been talking about reference for a long time so we have a lot of good material to draw from.

\paragraph{Frege: Sense and Reference} 

Consider the following example from Gottlob \citet{Frege1892-oh} (``Über Sinn und Bedeutung'' / ``On sense and reference'').:

\ex. \label{ex:frege_back} The morning star is the evening star.

Where \emph{the morning star} and \emph{the evening star} are two different referring expressions with distinct meanings, but they both in fact refer to the same object, namely the planet Venus. That is, the two expressions refer to the same thing, though they are expressed differently. 

It is important to understand that, though we use words and phrases to refer to things, the thing they refer to isn't the meaning of the words and phrases. To distinguish, Frege explained the difference between \emph{sense}, what we would call the meaning or notion of a word or expression, and the \emph{reference}--that is, the referred entity itself. The entity itself isn't the meaning, rather it instantiates something to which the sense can refer. 

\paragraph{Mill: Connotation and Denotation}

Before Frege, \citet{Mill1846-yr} made a similar distinction between \emph{connotation}, the properties or attributes that are implied by a word or expression, and \emph{denotation}, what an expression applies to in the world (i.e., the referred entity). This distinction is illustrated in Example \ref{ex:nps}:

\ex. \label{ex:nps}
\a. J: Did you hear that Sarah has a dog? \label{ex:nps_a}
\b. K: Yes, I was there when she bought it.\label{ex:nps_b}
\c. J: Ah, so the dog is real. \label{ex:nps_c}
\d. K: Yes, yes, her dog's name is Biff. \label{ex:nps_d}

where in \ref{ex:nps_a} no particular dog is being referred; the usage of the word \emph{dog} connotes a type of entity that has properties belonging to dogs, and \ref{ex:nps_b} where K is denoting a particular dog (which has all of the properties that the word \emph{dog} connotes). In this way, \textit{a dog} isn't really being used as a referring expression to a specific dog, but abstractly as a possibility, then J comes to learn that there is a dog that they are referring to (denoting). 

\paragraph{Wittgenstein} 

Philosophers have had a lot more to say about language than just referring to things, and no primer on language is complete without referring to Wittgenstein. Researchers often cite Wittgenstein for \emph{language is use in context} and \emph{language games}. What do those mean?

We defined language as a means of \emph{symbolic communication that occurs naturally within a human community}, with the important properties of \emph{by process of use, repetition, and change}. Those properties of, repetition, and change are what Wittgenstein is pointing to: language isn't just a static set of facts that we refer to; rather, it is something that is used much like a tool, but the tool itself changes itself (language is a \emph{process}). 

Language games are how language is instantiated and used in the real world. For example, speaking on the phone is one kind of language game, as is speaking with a bookstore employee who is giving me suggestions for certain genres of books and who they might be appropriate for. The words we use (and do not use), how we ask for information, how we give information, are all part of a specific language game.

\subsection{Compositionality} 

When we use words to communicate with each other, we don't use words in isolation. We use words in the context of other words. Individual words have meaning, but so do phrases, sentences, and paragraphs. \emph{The meaning of a complex expression (such as a sentence or document) is determined by the structure and meanings of its constituents}, an adage that is known as the \emph{principle of compositionality} \cite{Szabo2020-fd}. That is, a phrase is composed of words, sentences are composed of phrases, paragraphs are composed of sentences, etc. 

Each sentence in the above example dialogue \ref{ex:nps} about Sarah's dog only makes sense to a reader if they know the meanings of each word (including their senses and in most cases their references) and English syntax (the structure) to combine the meanings of words into a sentential meaning, and the even more complex meaning that is the entire dialogue. That's compositionality. 

We take compositionality for granted when we speak, but it remains somewhat unclear how it all works. Clearly grammar/syntax plays a role, but even if a sentence is completely grammatical, it might be nonsensical. Chomsky's famous example \emph{colorless green ideas sleep furiously} is a perfectly grammatical sentence, but there really isn't a way to compose it into a meaningful sentence. How are meanings composed into a greater whole? That's a big challenge for computers, though recent models seem to do something about it. 

In the sections that follow, we look at theories of semantics that are actually used in computers. We begin with formal semantics, then look at grounded semantics and distributional semantics. 

\section{Formal Semantics}

How can computers process and understand language? How can we can encode and represent linguistic meaning on computational devices? Since we ultimately want computers to process language, why not start with the closest thing we have to how computers compute? \emph{Formal semantics} looks at language in a strikingly similar way that computation works: logic. Logic is the study of correct reasoning and computation is concerned with well-defined calculations.

\subsection{Computation and Logic}

Computers operate on 1s and 0s. While 1s and 0s look a lot like numbers, the logical analog is True or False. If you open up your computer and look at the microchips, you'll see some green boards and metal parts. If you were to look closer--much, much closer--you'd see little things that are called logic gates. The idea is if we pass 1s and 0s (i.e., \emph{bits}) through the gates together, they will be able to make a comparison and output the result. Some examples:

\ex. \label{ex:logic}  
\a. 1 AND 1 = 1
\b. 0 AND 1 = 0
\c. 0 AND 0 = 0
\d. 1 OR  1 = 1
\e. 1 OR  0 = 0
\f. NOT 0 = 1

So if we have an AND operator and give it a 0 and a 1, it tells us 0. If we give it a 1 and a 1, it tells us 1. That doesn't seem useful, but connect these things together in different ways and you get computation including memory management, processors, and beyond. Simple, yet elegant. There are others, but the AND, OR, and NOT operators an get us pretty far. A computer has many millions of these gates stacked together and can use those gates to move data around, perform complex computations really quickly, and do it all using electricity instead of something more costly. 

It makes a lot of sense to use the ``native language" of the computer (i.e., logic) and try to use that as a basis for representing human language, then we can use the machinery of the computer to do all of the processing, at least that's what the theory states. The challenge is mapping from the things we say and write to a representation in logic that reflects them. That's where the study of formal semantics comes in, beginning with First Order Logic. 

\subsection{First Order Logic}

First order logic (\fol) and research into other logics has a long history that has influenced linguistics as well as the architecture of computers.\footnote{I refer the reader to \citet{Ewald2018-xv} for an overview of the history, and Chapter 2, section 3 of \citet{Kennington2016-pl} for an overview of other logics that have been applied to the study of semantics.} \fol\ defines operators like AND and OR, things that the operators operate on which looks like a reasonable fit for, respectively, structure and meaning. 

For example, if we replace 1s and 0s with words, we can relate words to each other using the logical operators:

\ex. \label{ex:fol}  
\a. big AND gray AND elephant = the big gray elephant
\b. taco OR salad = I want either the taco or the salad
\c. NOT here = John is not here

The task, then, is to with take statements and translate them into their corresponding logical representations. A simple example for the sentence ``the gray thing":

\ex. \label{ex:fol_small}  
\a. $\exists x. gray(x)$

Think of gray as a function and $x$ is being passed into it, and the function has to return either True or False. The backwards E means ``there exists" (in this case, a thing). If nothing can be assigned to $x$, then the statement is false. That is, if there isn't a gray thing, then the meaning of the statement is that it is false.

We can then combine these using logical operations, which is great because language can be very complex. An example of a logical representation for more complex language: 

\ex. \label{ex:fol_big}  
\a. $\exists x. big(x) \wedge gray(x) \wedge elephant(x)$

In this example the variable $x$ has to simultaneous be big, gray, and an elephant to be True. The point of being true is an important one here: what we are asking of this formula is to assign $x$ to something that fits all of the three requirements at the same time.

\subsection{Shortcomings with Logical Forms}

The thing about big, gray elephants is that they are physical objects in the real world. So, how does the logical system know about what is big, what is gray, and what is an elephant? There's an entire theory of logic and special notation for that, too. This is where we get into set theory and modal logics. I recommend the interested reader to learn more about it or take a course on logic. If anything, such a study helps a person understand common logical fallacies that happen all around us. 

Gray, big, and elephant are physical characteristics of entities in the real world. Sure, we can talk about them and represent what we say about them as some kind of logical form, but if we are talking about semantics---the meaning of words and more composed language---how does the machine know what \emph{gray} actually means? How does it know when \emph{gray} should return true, given some variable? That wasn't being solved by \fol. 

We now look at another way to view semantics that attempts to uncover and address these shortcomings. 

\begin{figure*}
\centering
   \includegraphics[width=0.95\linewidth]{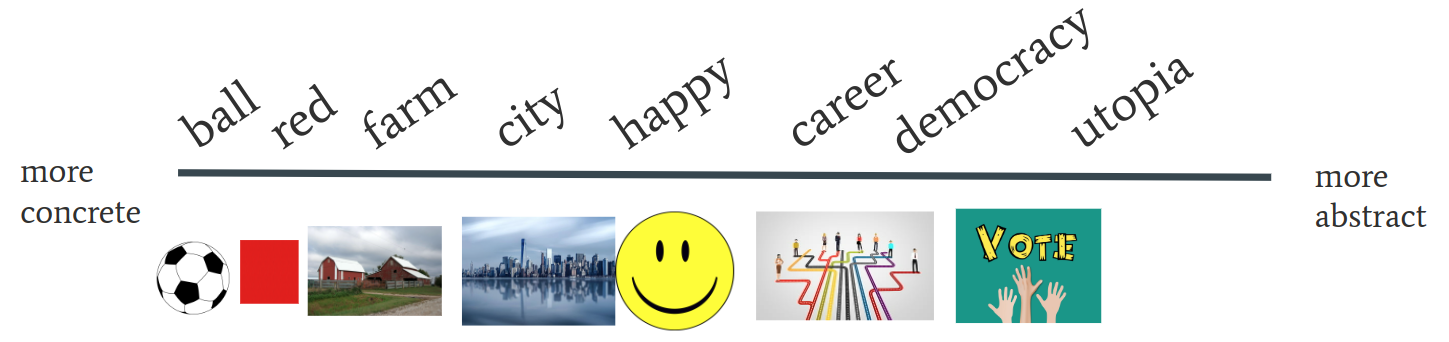}
  \caption{Examples of words that are more concrete vs. more abstract. Words that are concrete have physical (in this case, visual) denotations, whereas more abstract words do not physically exist. Concreteness ratings from \citet{Brysbaert2014-jc} resulted in the placement of the words. Figure borrowed from \citet{Kennington2022-db}.}
  \label{fig:concrabst}
\end{figure*}


\section{Grounded Semantics\footnote{Some of the ideas and wording for this section come from my prior work \cite{Kennington2022-db}.}}

One criticism of logical approaches to meaning is: where is the meaning? If we consider what meanings of words (or phrases, etc.) are, they are symbolic representations of something else. That doesn't seem to be a problem given the definition of language above (``a means of symbolic communication"), so all we need are symbols (words or even logical representations), right? 

In his famous 1990 paper \cite{Harnad1990-fr}, Stevan Harnad explores this question. Harnad pointed to \citet{Searle1980-wc}'s Chinese Room as a metaphor which challenges the core assumptions that symbols carry meaning on their own. He explains that if he, someone who could not read or speak Chinese, were in a room with a Chinese-Chinese dictionary and had the instructions to take ``input" of one Chinese character, look up the character in the dictionary and then find the ``output" character, even if the inputs and outputs were perfectly mapped as observed by an outsider, the person in the room doesn't actually know Chinese, which is the same problem that computers have when they process natural human language, even for formal representations like \fol.

Yet are words not also symbols? In some ways yes (as we defined symbols above), but we need to be clear here what is meant by \emph{word}. A word is a linguistic unit that carries linguistic meaning on its own and can be used as a placeholder for a concept much like symbols can. For example the word \emph{chair} can denote real chairs, but uttering or writing the word can replace the presence of chairs when someone wishes to talk about the connotation of a chair---the word \emph{chair} effectively becomes an abstraction of the connotation. The confusion comes when one assumes that the word \emph{chair} as it is written actually represents the connotation itself, but it does not. The connotation of \emph{chair} resides in human brains, but because written text is computable and since text is a placeholder for concepts for humans as they communicate with each other, it follows that machines could use text as symbols and text would carry the meaning. However, that is precisely what the Symbol Grounding Problem is pointing out does not work because, like symbols, text is ungrounded.

What do we mean, then, by \textit{grounded}? Simply put: the meaning of many words is found in our experience with (i.e., \emph{grounded into}) the world. We know what chairs are, not because we've read about them, but because we've experienced them. We have seen them so we know what they look like, and we have used them for sitting so we have muscle memory of what it means to fit into them, and we have felt the relief of sitting in a chair after spending a lot of time on our feet. All of those things are part of what what the symbol \emph{chair} means to us.

Word meanings can ground into all sensory input. \emph{Red} grounds into vision, \emph{smoky} grounds into smell, \emph{sharp} grounds into touch, etc. Beyond sensory inputs are other internal modalities, such as haptics and muscle memory. For example, you could close your eyes right now and make a "thumb's up" gesture with your hand because you have grounded that into the muscle memory of what that gesture feels like--you don't need to look at your hand at all. Moreover, verbs like \emph{kick}, \emph{swallow}, \emph{wave} are all things you can do by muscle memory; in fact, you likely could do them before you knew the words for them. That's symbol grounding.

\subsection{Concrete and Abstract Meaning}

Concrete words are words that denote physical things like objects, shape, and color (e.g., \emph{chair}, \emph{red}), requiring Symbol Grounding to arrive at meaning, whereas abstract words are words that denote ideas (e.g., \emph{democracy}, \emph{travel}) that are often defined by other words. It should be noted that the distinction between concrete and abstract concepts lies on a continuum, not a binary dichotomy \cite{Della_Rosa2010-th,Brysbaert2014-jc}. Thus some words are more concrete or abstract than others, some examples that illustrate this are shown in Figure~\ref{fig:concrabst}. Words range from very concrete (e.g., \emph{ball}) to very abstract (e.g., \emph{utopia}). For more concrete words, corresponding images show clear examples of something that the word can denote visually. However, more abstract words can have aspects of their meaning represented visually, but not fully (e.g., \emph{democracy} includes voting, but voting is only one aspect of the meaning of \emph{democracy}). 


That some words need grounding while others do not begs the question \emph{Which words need symbol grounding?} Words that are more concrete like \emph{ball} and \emph{red} clearly need to be grounded to be meaningful. The word \emph{red}, for example, can be understood to some degree without grounding, for example that it is a color and that certain objects can be red (e.g., apples and vehicles), and while it is true that there are metaphorical uses for the word \emph{red}, those metaphorical uses can only be understood after knowledge about \emph{red} as a color is learned (see arguments made in \citet{Lakoff2008-ud} about metaphors; see also \citet{Bizzoni_undated-on} for discussion on visually grounded metaphors). More recent work has shown that all words, including abstract ones, ground into something \cite{Banks2023-qn}.  

On the other end of the continuum are abstract words like \emph{democracy} and \emph{utopia}. Even though someone could imagine a visual depiction of either of those terms, their meaning is not grounded directly into the physical world, but are rather ideas that are defined by other words. 

\begin{figure*}
\centering
   \includegraphics[width=0.9\linewidth]{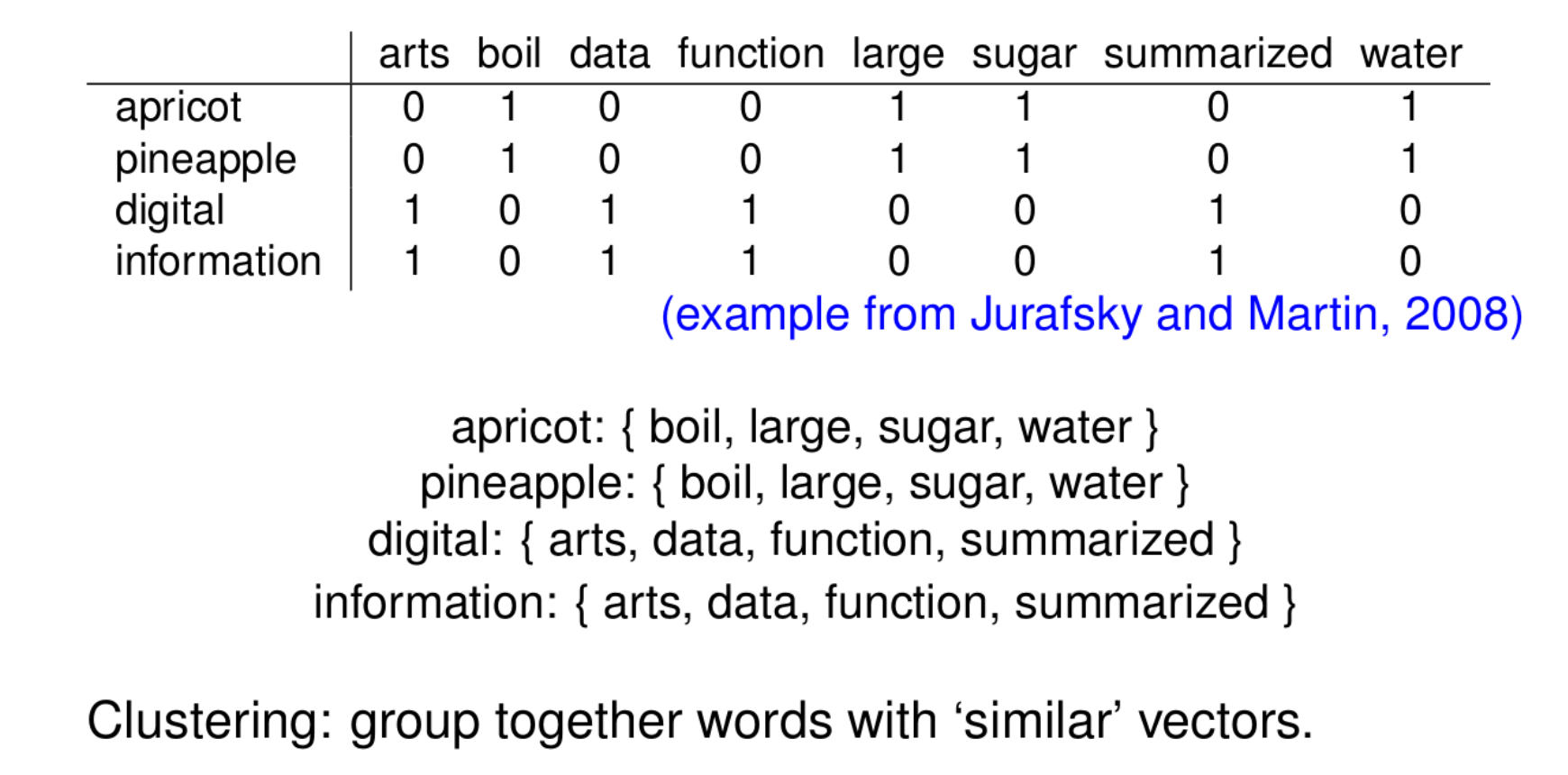}
  \caption{Example of distributional semantics by counting words in lexical context. Borrowed from \cite{Jurafsky2024-qx}.}
  \label{fig:distsem}
\end{figure*}



\subsection{Top-down vs. bottom-up Language Processing}

Grounded semantics makes a strong case that much of language learning, particularly meaning, happens in a bottom-up fashion where concrete words are learned through interaction with the objects they denote in the real world, and those words bootstrap the learning of more abstract words. For example, after a child learns words like \emph{red}, \emph{green}, and \emph{yellow} can be abstracted into the category \emph{color}. However, it is clear that there is much top-down processing in language processing. When two people are speaking to each other, a listener is predicting top-down the tone of the speaker \cite{Shuai2014-si}, and it is well-known that humans predict the syntactic categories of words yet to be spoken. 

\subsection{Shortcomings of Grounded Theories} That grounding should take place is clear, but how a machine should ground is not so clear. Many researchers claim to address symbol grounding in their papers. It's true that some models can make use of visual representations (like convolutional neural networks) to identify objects and object properties which has a form of groundedness, but how those representations are combined into a language model of some kind seem arbitrarty based on technical constraints---not based on what it might mean for the semantic representation of the model. Machine learning and deep learning classifiers have opened many doors to grounding symbols into more fine-grained representations with varying degrees of success. Vision is the most represented modality to ground into, but so many others exist including but not limited to olfactory, tactile, haptic, vestibular, interoception, and muscle memory. It is easy to see progress on bringing the physical world into models of language, but in some cases it could be that they fall victim to the symbol grounding problem.

\section{Distributional Semantics}

Who hasn't found themselves reading something in an article or book and come across a word that they had never seen before, but given the context of the words and sentences around it, were able to at least get an of what the meaning of that word could be? Firth's idea \emph{you know a word by the company it keeps} makes intuitive sense, otherwise one wouldn't really be able to read and understand language as it is written \cite{Firth1957-xy}. The point of written language is to communicate symbolically using a medium other than speech, and we clearly learn some words by how they are used in text. Could we somehow model that process computationally?

\textit{Distributional semantics} posits that the meanings of words can be estimated based on how they are used within language itself. It is important to note here that language largely refers to written text, and using text is helpful because (1) it is easy to find all over the Internet and (2) it is much easier to process on computers than speech. 

\subsection{Early Distributional Theories}

The challenge is that text on a computer is not much different from text on the page of a book; on its own, it's a meaningless, symbolic representation. Words represented by text don't have any intrinsic meaning; meaning is brought to mind as we read them because we know what the words connote and denote. So we are back to the original problem of computational semantics: how to represent the meaning of a word in such a way that computers can process them. For \fol, that means symbolic logic, but computers are also really good at processing numbers. Can we somehow use numbers to estimate meaning in some way?

We have some options here: we could represent each word as a number. For example $the=1$, $a=2$, ..., $kite=477$, ..., $chair=533$ and so on. Does that really get us anywhere? How can we use words represented as numbers to estimate some kind of meaning? One answer might be: words that have similar meanings should be closer to each other, and words that have different meanings should be far from each other. That makes intuitive sense, but it is complicated because words like \emph{kite} relates to \emph{wind} as well as \emph{park}, but then \emph{wind} is a word that relates to other weather phenomena that aren't important for kites or parks. We need another way to represent words using numbers. Can we use more numbers than just one? Yes. 

We can take the vocabulary of a language (i.e., the unique words), and treat each word as a list of numbers equal to the length of the vocabulary. All lists have mostly 0s, but they have a 1 in a unique place. For example, $the=[1,0,0,...]$ and $a=[0,1,0,...]$ and so on. Those lists of numbers are called \emph{embeddings}, or, more technically, \textit{vectors}, and these vectors that are all 0s except for a single 1 are called \textit{one-hot} vectors because they are ``hot" in one place. A vector is a list of numbers, whereas an embedding means that the vectors are not just distinct from each other--they are related. More precisely, vectors/embeddings are points in the same n-dimensional space. 

What's really nice about vectors is that we can use some very well-developed mathematical machinery used in the field of linear algebra to work with vectors. We can, for example, multiply, add, subtract, and find distance between vectors in different ways. One nice thing about one-hot vectors is that are made up of all 0s except for a 1 in a unique place (i.e., each word is a \textit{unit vector} because it only has a length of 1) for each word means that words are all equidistant to each other word, so there are no assumptions about distance and word meanings since all words are equally spread apart from each other. This solves the problem of numbering words to represent their meanings and this gets us started representing words as lists of numbers known as vector embeddings. However, what we really want are smaller vectors (because big vectors are hard to work with computationally) and we want the vectors to have more information in them such that words that have similar meanings are closer to each other, not equidistant.

One idea is to take a big corpus of text, say Wikipedia, and find the vocabulary. Then we can pick, say, the top 1000 most common words used in Wikipedia and use those as a basis for comparison. Then, for every other word in the corpus, we find each time it is used and look at the 5 words before it and the 5 words after it (the company a word keeps). If any of those words are in the list of 1000 top words, we count them up. 

The top of Figure~\ref{fig:distsem} shows a simple example of this idea. The words along the top (\emph{arts} to \emph{water}) are the most common words and the words on the left (\emph{apricot, pineapple, digital, information}) are the words for which we are trying to find the right vectors. After counting, we find that \emph{apricot} and \emph{pineapple} show up around words like \emph{boil, large, sugar, water} whereas \emph{digital} and \textit{information} show up around words like \textit{arts, data, function} and \textit{summarized}. If we look across each row, we see 1s and 0s, and if we squint a bit we can see that \textit{apricot} has a similar row to \textit{pineapple}, and \textit{digital} has a similar row to \textit{information}, yet the fruits have different rows from the other two words. Now we have more than just a single number representing words, those numbers were found by only using text, and words that have similar meanings seem to have similar lists of numbers--i.e., vectors!

Now we have a basis for a semantic theory: words that have similar vector embeddings (i.e., points) have similar meanings, whereas words that have different embeddings have different meanings. The distance between them correlates with how similar or different the meanings of words are from each other. 

While vectors/embeddings are nice because we can represent meaning using numbers which computers can process easily \cite{Turney2010-pc}, early embeddings found by just counting words in their contexts were too big to efficiently process, and they were too \emph{sparse}, meaning the vectors didn't have very much information in them, often due to how many 0s there were in each vector. If only there were a way to package more information into smaller vectors. 

\subsection{word2vec}

\citet{Mikolov2015-it} took the theory that we could model word meaning by deriving embeddings from how words are used in conjunction with other words to the next level. Instead of simply counting words in a corpus of data, they modeled the embeddings directly. They were able to take the huge one-hot vectors of all 0s except for a 1 in a unique place, and shape them into vectors that were smaller and now there were no more 0s: each vector had numbers between -2 and 2 with precise decimals. These vectors were easy to train on text and were more useful to computers than anything before. Effectively, their word2vec model played a straight-forward game of guess-the-word given the words around it, resulting in a smaller vector that was more manageable. 

Suddenly, smaller vectors with more information were being used for everything in \nlp. Smaller vectors (usually somewhere between 100-500 dimensions) are amenable to machine learning classifiers that love numbers as input. Word2vec embeddings helped improve machine translation models, sentiment classifiers, among many other applications where text can be processed. 

Despite improved representations of vectorized words (e.g., GloVe \cite{Pennington2014-tv}) distributional representations had their limitations. As Raymond Mooney said in his semantic parsing workshop talk in 2014, "You can't cram the meaning of a whole $!@\%@$ sentence into a single $@!\%!@$ vector!" And that was the problem: the meaning of the vectors was at the word-level. How does one put the words together to form the meaning of an entire sentence? There are operations we can take on vectors, but using those did not seem to do the trick of actually composing the meanings of sentences from their constituent words. Adding vectors together, for example, did not preserve any notion of syntax or word order in the final vector. For example, what does it actually mean when we add the vector for \textit{red} and \textit{car} to arrive at a vector for \textit{red car}? Some clever work showed that representing different word types using different structures (i.e., some words are vectors, but others could be matrices---a vector of vectors) could preserve some of the composition \cite{Baroni2010-vh}, but the methods didn't scale beyond two-word sequences. Phrases and sentences can be arbitrarily long (the average English sentence length in Wikipedia is 25 words).  

Can we take the idea of distributional semantics beyond the word level? The answer to that question came in 2017 and it changed everything. 

\subsection{Transformer Language Models}

Suppose I begin a sentence and ask you to continue it: \emph{I want a scoop of ...}, you would probably say something like \emph{ice cream} (I was thinking guacamole, but both work) because we associate getting ice cream in scoops, and many people want ice cream, so seeing words like \emph{want} and \emph{scoop} lead us to ice cream. Predicting what comes next based on what is already seen is the basis of how \emph{Language Models} (\lm s) work. The first \lm s did something like we looked at above with distributional semantic vectors: they just counted how words followed other words and kept track of the statistics. Then we could use \lm s for useful things like machine translation of automatic speech recognition because a language model could provide the probability of a sentence being a "good" one based on data.\footnote{My master's thesis resulted in a language model that could have (theoretically) infinite context \cite{Kennington2012-vi} based on words-that-follow-other-words statistics, and it helped improve machine translation at the time.} Language models had their uses over the years, but were somewhat forgotten when word2vec was introduced because language models only captured statistics of word sequences, not meaning. 

That all changed when \citet{Vaswani2017-kv} became public. That paper introduced the \textit{transformer}, which takes inspiration from language modeling in how it is trained, and from word2vec in how words are represented. Similar to word2vec, transformers are trained with large amounts of text, words are predicted based on the words around them in the text, and the job of the model is to guess the word. The models went beyond the word level, however, in that the vector embeddings it produced were not just word embeddings, but word, phrase, sentence, and even document-level embeddings. Somehow, words are first represented at the word level, but then composition happens by vector and matrix manipulations using intricate and clever linear algebra that we won't go into here, operations which deep neural networks facilitate automatic learning, given enough data.

Transformer Language Models---the underlying models used in large language models (LLMs) like ChatGPT and others---did two really important things at once. First, the authors showed that the model could be trained on large amounts of text once, then they could be used for any text-related \nlp\ task such as translation or sentiment classification. This \textit{pre-train} on text then \textit{fine-tune} on a specific task changed the paradigm forever because we could take a pre-trained model and fine-tune it for our specific needs. Second, the authors showed that a single model could do almost anything \nlp-related. Whereas before, researchers could spend an entire career focusing on a narrow \nlp\ task like translating from German to French, suddenly, in one day, one model performed better on multiple benchmarks on many \nlp\ tasks all at once. Within months, \nlp-related conferences were seeing more and more transformer-based \lm s in just about every paper. By 2022, the models had become sophisticated and scaled enough to form the underlying architecture for models like ChatGPT. Suddenly words like "GPT" and "language model" were not just spoken in \nlp\ circles, but by everyone. 

\subsection{Shortcomings of Distributional Semantics}

It has been argued that transformer-based language models are not distributional because they go beyond co-occurance counts that the original distributional methods used. That's true, but distributional doesn't just mean co-occurance; it means that the meaning of words can be derived from how they are used within text. Co-occurance counting, word-level vector embeddings, and even transformer based language models which learn based on guessing words, are all in my opinion distributional in nature. 

Distributional models have really improved since word2vec in 2013, and form the basis for most of the chatbots that are widely used today. Clearly, the transformer architecture has been successful and transformed the field of \nlp. However, it should be stated very clearly again that \emph{text is not language}, and text certainly is not meaning. Text is a way of representing language, but language is written by language users, read and understood by language users, and those language uses know meanings of the words they write and read. 

A distributional model like an LLM that is trained only on text has no notion of the concrete meaning of words. The first thing I asked ChatGPT when it came out was \emph{Have you ever seen an apple?} It answered that it has never seen anything, let alone an apple. Apples have meaning because we experience them physically, we know how they look and feel, and how they taste when we eat them. Someone might love a drink or food that is derived from apples, and therefore has deeper meaning. Others might make a living by growing and selling apples. Thus apples mean something \textit{to us}, but apples don't mean much \textit{to models}. Vision language models are able to model meaning from images and text which is a step in the right direction (see \citet{Fields2023-zv} for an overview), but images are only a small portion of our physical experience where we derive meaning and connect that meaning with language. 

\paragraph{Wittgenstein and Grounding} \citet{Kennington2022-db} pointed out that \citet{Wittgenstein2010-ji} sometimes brings up color and shape (1.72-74) and that words refer to objects. Could Wittgenstein have meant that \emph{context} is not [\emph{just}] lexical context, but physical context (or some degree of both)? This is an important question because Wittgenstein (along with Firth) has always been called on to motivate distributional methods of language modeling, yet words keep company with more than just other words, including words that are more concrete." which also points to grounded semantics. 

Clearly a lot of useful information, even a degree of linguistic meaning, can be derived from text. However, human children do not learn their first language through the medium of text. Does it matter that computational models like transformer-based language models learn differently from humans? In the next section, we explore what is known about how human children learn language to consider the differences between human language acquisition and how language models learn language.

\section{Human First Language Acquisition} \label{sec:first-language-acq}

In this section, we survey some of what is known how children learn language then compare/contrast that to how computational models, including language models, learn language.\footnote{Some of the content for this section is taken from \citet{Kennington2023-ft}.} This survey is meant to point out potential aspects of language and semantics that might be important for a model of computational semantics. 

\subsection{Child and Human Development}

The field of child development is a sub-field of psychology, but also biology and sociology.\footnote{Anyone venturing into first language acquisition should refer to \citet{Clark2013-rs}.} Children begin to speak their first words fairly early in life (about 12 months), despite the amount of language that they are exposed to being very small (a few million words is the current estimate). The most fundamental and natural way for humans to communicate with each other is interactive, spoken dialogue \cite{Fillmore1981-lu}. According to \citet{Clark1996-zu}, to learn a language a child learning a language must:

\begin{itemize}
    \item be situated (speakers and listeners must be in the same shared space)
    \item have shared attention (speakers and listeners must be able to see what other people are pointing or looking at)
    \item use speech as the primary medium
    \item agree on how words are used to refer 
\end{itemize}

Jean Piaget is well-known for his theory of childhood development which has four stages: (1) sensorimotor sage (birth to 18 months) when infants begin higher-order mental activity (e.g., reasoning and language), (2) Pre-operational stage (2-7 years) when children can begin to consider concepts that aren't directly in front of them, (3) Concrete operational stage (7-11 years) when children can mentally simulate situations without actually playing them out, and (4) Formal operation stage (12-15 years) where children can think more abstractly and test hypotheses using deductive reasoning. Some of the stages have sub-stages. 

Child development is important to language because at all stages, children are able to perceive and operate on the world in different ways which alters their language comprehension and production abilities. Moreover, language is part of how children interact with others, organize their understanding of the world, and foster relationships with others \cite{Alan_Sroufe2009-wa}. The study of child development, particularly of how children learn their first language, is a field of study that all other fields that deal with language should draw inspiration from. 

\subsection{The Setting: Situated, Spoken Dialogue}

In his seminal book, Herbert Clark explains the most basic setting for language use, which the setting where children first learn their language \cite{Clark1996-zu}:

\begin{itemize}
    \item situated - multiple people can directly perceive each other in a co-located situation in time and space
    \item shared attention - people can use extra-linguistic knowledge to communicate including pointing gestures, and visual saliency directs the attention of language users
    \item speech - the primary medium that people use to communicate is speech (children cannot yet read and write, and many languages do not have writing systems); alternatively, signed language is also primary, but speech is more common
    \item joint activities - people use and hear language and update their understanding of language through experience of language in activities with other people
\end{itemize}

The basic setting for language is spoken interaction. It may not be the most common cite of language use (perhaps SMS texting, emails, or other mediums are more common in industrialized countries), but it is the most basic. Children first learn to speak (or sign) before they learn literacy; i.e., reading and writing. 

\subsection{Attention \& Joint Attention}

\citet{Clark1996-zu} brings attention into the language use (and learning process) because without attention, we likely wouldn't be able to learn language at all. 

We as humans can only attend to one thing at a time, and even though we have multiple senses, we tend to filter out things that are not within our current frame of attention. For example, if you are listening to someone on the phone in one ear and someone tries to tell you something in your other ear, you can't attend to both at the same time. The same happens in vision: even if we look out on a scene that is full of many objects and actions, such as people walking across a wooded campus while the sun rises, we can only attend to one narrow area at once. Attention can be so focused as to filter out very novel things. For example, in a study, researchers asked participants to watch a video and count how many times a team of people passed a basketball to each other. After several minutes of counting the researchers asked the participants \textit{did you see the gorilla?} Sure enough, when shown the video again the participants could easily see the person in the gorilla suit, yet many missed it the first time because their attention was so narrowly focused on the goal of counting basketball passes \cite{Baggio2018-lp}. Others have refuted the work do some degree \cite{Wallisch2023-se}, but none refute the importance of attention. 

Attention is as critical to language as it is to human cognition. When children learn their first words, it is often due to their attention being focused on one thing. If a child holds an object and looks at it, the caregiver will often say the word for the object instead of a word for an object that is in the adjacent room or even in the same room as the child but the child is not focused on it. Caregivers know intrinsically that attention on an object is a precursor to assuming that the object is what a word refers to. When the caregiver and child are both attending to an object, and both know that the other is attending to an object, this is known as \textit{joint attention} and is necessary for the first words that children learn. 

Later, the caregiver can refer to objects that are not in the child's attention \textit{in order to draw their attention to the object}. For example, the caregiver knows that the child has learned the word \textit{spoon} in a prior interaction and uses the word \textit{spoon}. The child responds by looking for a spoon, picking it up, and handing it to the caregiver. 

\subsection{Referring to Objects}

Among children’s earliest communicative attempts are acts to indicate objects for other people, for example, pointing to an object or holding up an object to show it \cite{Wittek2005-uu}. Once language begins, children rapidly acquire a host of additional linguistic capabilities (see \citet{Piaget1951-an}) including learning how words not just denote physical objects, but also actions (i.e., verbs) and words can be strung together in more complex phrases and sentences. Moreover, Children who are learning their first words learn words slowly \cite{Westermann2017-da} and there is a strong correlation between speed at which words are learned and how much parents talk to their children. 

Even though there isn't a specific curriculum that caregivers administer to kids to get them to talk, there seems to be some patterns. \citet{Hetherington1999-lv} found that parents repeat what small children say, they take very clear dialogue turns, and caregivers often rephrase what kids say in a grammatically correct way (and correct pronunciation). The known \textit{zone of proximal development} seems to be intrinsic to mothers who speak with their children: they keep a level of complexity just ahead of the child which gives the child novelty as well as comprehension. 



\subsubsection{Incremental Processing}

When two people are conversing with each other, they take turns being speaker and listener and responding in real-time. This real-time constraint on language generation and comprehension is important: because of the limitations that humans have to attend to one thing at a time, and because speech is conveyed via compressed air waves between the speaker and the listener, speech must be produced syllable by syllable, word by word, over time. Humans are unable to transmit large chunks of information all at once so a full phrase, sentence, or paragraph cannot be somehow signaled from one person to another. One could argue that a text SMS or email can do just that, but the writing of the text/email and the reading of it later by the recipient must both be done \textit{incrementally}, i.e., word-by-word. Indeed, \citet{Tanenhaus1995-rb} showed that speech comprehension happens at a word or sub-word level. 

The idea that humans produce and comprehend spoken (and even written) language incrementally from childhood until their last utterance seems obvious, but it needs to be pointed out here because most dialogue systems and chatbots \textit{do not} process incrementally. Language models generate language one word at a time, but the underlying architecture (e.g., the transformer) is designed to process many words in a sentence or paragraph in parallel---most assuredly not incrementally. Does that matter? Perhaps not, but it could be argued that something is inherently wrong with a model of language understanding that does not process in the same way that humans do because language is such a human capacity.\footnote{See arguments made in \citet{Kennington2025-ih} which gives a fairly detailed overview of incremental processin in automated systems.}

\subsubsection{Building Common Ground: Clarifications \& Conversational Grounding} 

Language is not so much of a thing, but a process. Humans acquire language (in many cases, multiple languages) throughout our lives. We learn many words during our years of formal education and, as noted above, those words move from concrete to abstract.

Not only do we learn new words throughout our lives, we also gain a deeper understanding of what words mean. For example, someone who has experienced cancer either personally or in a loved one sees that word more than an abstract concept that has a definition. Someone who has never physically seen a zebra might know that they look like a horse with black and white stripes, but the level of understanding is different from someone who has seen a zebra directly. 

Not only do we learn new words, and not only does our understanding of words change throughout life, the way we use and understand words can change throughout the course of a conversation. Imagine two people in a cafe having a conversation. Person A says \textit{I was feeling pretty melancholy yesterday} then describes her experience as really keeping her from accomplishing anything beyond just getting through the day. Person B listening to this had thought before their conversation that the word \textit{melancholy} was perhaps not so drastic as to be synonymous with a feeling of depression, so Person B updates their understanding of how the word can be used. 

This give-and-take of use and update of understanding is known as \textit{conversational grounding}, also explained in \citet{Clark1996-zu}. When we use language, we communicate about events, feelings, plans, goals, etc. But often we come across speech events that require us to make repairs. We often ask people to repeat something they said because we didn't hear them, we didn't understand a reference or a word, or there was some other ambiguity. These requests for repetitions are a form of \emph{clarification request}, and we use them all the time. 

\citet{Purver2004-jc} showed that from all domains in a corpus of transcribed text (the British National Corpus \cite{Lou2000-lz}), around 3.5\% of dialogue turns (418/11,800) had some kind of communication breakdown which resulted in a clarification requests.\footnote{See also \citet{Ginzburg2012-wm}, Chapter 6 for a detailed analysis on clarification request types.} In spontaneous dialogue between two people, there is a higher degree of communication breakdown, between 3-6\% \cite{Rodriguez2004-nk}, necessitating a need for effective clarification strategies to mitigate breakdowns in communication. 


Conversational grounding is distinct from symbol grounding, but one can act as scaffolding for the other \cite{Larsson2018-qo}. As two people interact and learn about how words are used (conversational grounding), one of the conversation participants could point to a flower and say \textit{that's a Dahlia}, giving the listener a new word and visual knowledge about what the word denotes (symbol grounding). 

\subsection{Affordances}

Important to our understanding of objects is how they look: their shapes, colors, etc., but also important is what they can be used for. A chair, for example, has a shape and a particular chair may have a specific color or two, but what is important to humans about chairs is their use: we can sit on them. 

\citet{Gibson1966-gk} introduced the term \textit{affordance} to conceptualize the fact that when humans look at other things, they look beyond just surface structure and infer \textit{what the object can do}. Chairs are for sitting. Brooms are for sweeping. Ladders are for climbing. Balls are for kicking or throwing. Spoons are for eating. Etc., etc. 

Lingusitic meaning has a lot to do with affordance. Meaning often is not an intrinsic property of something, but rather what the thing \textit{means to us}. What does a chair mean? It isn't just an object, it means that I can rest after a lot of standing and walking. What does a glass of water mean? It means I can quench my thirst. The way an object is meaningful to us is directly tied to what kinds of actions we can take on the object. 

It has been shown that language models can acquire knowledge about affordances of objects, mostly because those affordances are talked about in the text that is used to train language models \cite{Forbes2019-sq}. Does that mean that a language model can learn what is meaningful to humans? Most likely, as long as someone wrote it in some text somewhere that is used to train the model. Does that mean that objects like chairs and glasses of water are meaningful to language models? What is meaningful to language models is an open and important question.

\subsection{Exploration \& Curiosity}

Human children are curious, and often children exhibit their curiosity through play. Play means \textit{to engage in an activity for enjoyment rather than for a practical purpose}. Play happens early, even between infants and their mothers \cite{Stern1974-sx}. Curiosity and play are ways that people explore their world. Children who are not yet attuned to opportunities and danger are intrinsically motivated by curiosity to explore and play. It could be argued that exploration and curiosity are necessary precursors to language learning. 

Computational models of curiosity and exploration was explored by Oudeyer and colleagues \cite{Oudeyer2005-yi,Oudeyer2007-av}. Though their goals were not language learning, their work modeled important precursors to language learning; how can a child learn language without exploration, and why would a child explore without curiosity? Modeling curiosity is a challenging problem because what is it about a child's environment that would enable them to curiously explore? Oudeyer's work used information theory in that a particular setting if the model has experienced something similar before, it is not curious, but if there is something novel, the model explores that new thing. But how to explore? That requires action, and knowing what kind of action to take (affordance). Small children move their bodies seemingly randomly at first, but then more controlled as they get older. That means the model that uses curiosity to learn must be able to act in the world; in the case of Oudeyer's work, they used a robot.\footnote{Our ongoing work attempts to build off of this to allow robots to explore a more complicated space and incorporate the visual world into the model \cite{Henry2024-vs}.}

What does exploration have to do with meaning of language? If an agent that is learning a language cannot explore the world it lives in, what can it know about the world that the language can ground into?

\subsection{Intention}

Whenever people do anything such as eat food, exercise, socialize, or most other activities, they do it because they \textit{intend to} (as distinct from \textit{want to}; sometimes people intend to do things they don't want to, like exercise). Intent is defined as \textit{choice with commitment} \cite{Cohen1990-dk}. Intentions follow four functional roles:\footnote{Following Bratman's philosophical basis for intention \cite{Bratman1987-zy}.}:

\begin{enumerate}
    \item intentions normally pose problems for the agent; the agent needs to determine a way to achieve them
    \item intentions provide a ``screen of a admissibility" for adopting other intentions
    \item agents ``track" the success of their attempts to achieve their intentions
    \item agents must distinguish between possible and actual events\footnote{See section 1.5 of \citet{Cohen1990-dk}.}
\end{enumerate}

Furthermore, given the above functional roles, if an agent \emph{intends} to achieve a possible outcome $p$, then:

\begin{itemize}
    \item the agent believes $p$ is possible
    \item the agent does not believe they will bring about $p$
    \item under certain conditions, the agent believes they will bring about $p$
    \item agents need not intend all the expected side-effects of their intentions
\end{itemize}

For example, if I am feeling hungry while I am sitting on the couch and doing nothing, I am $p$: \textit{motivated to eat something}, and I believe $p$ is possible. However, the condition of sitting and doing nothing will not bring about me eating something (I don't believe I will bring about $p$ in my current state of sitting on the couch), so I need to change what I am doing to bring about eating something. I decide to stand up, go to the fridge, and find something (I will bring about $p$). I don't have to worry about a meteor hitting earth just then which might inhibit me from eating something or I don't have to worry about what walking to the fridge might mean for someone who later wants to eat something from the fridge only to find I made it there first (a side effect). 

These kinds of intentional actions are a constant part of life. The fact that I am writing this sentence means I intend for someone to read it, and you reading means you intend to possibly learn something from what I have written. 

Understanding other people's intentions is something that children learn very young. \citet{Rekers2011-er} showed that toddlers are collaborative. For example, someone holding an armload of books trying (i.e., intending) to open cabinet door, but struggles to open the door because their arms are full. Children (but not chimps) can recognize the intention/goal of the other person and open the cabinet for them without either of them saying a word.

What does intent have to do with meaning of language? Agents that are intentional are agents that explore and act in the world. People have the intention to socialize and communicate with others, and that has to be done with language. Moreover, understanding what other people say means we understand their intentions to a certain degree. 

\subsection{Theory of Mind\footnote{Some of the content of this section comes from \citet{Kennington2022-lt}.}}

Defined broadly, human \emph{Theory of Mind} (\textsc{tom}) refers to the capability that people can recognize, represent, and make inferences about the desires, beliefs, and intentions (see above section about Intention) of other people \cite{Premack1978-am}. \textsc{tom} has been studied in child development, cognitive, and psychological literature (see an overview in \citet{Baron-Cohen1997-il}), and has recently been explored as an important aspect of interaction between people and machines. From the side of the humans, it is well known that humans mentally attribute anthropomorphic characteristics to machines---robots in particular---based on physical morphology and behavior in many different ways including sympathy and intelligence \cite{Novikova2017-jp}, emotional state \cite{McNeill2019-ia}, age \cite{Plane2018-wh}, gender stereotypes \cite{Eyssel2012-fp,Kuchenbrandt2014-vt}, and social group membership \cite{Eyssel2012-px}, which suggests that humans attempt to apply \textsc{tom} to a certain degree to other agents including humans, animals, and machines like language models and robots. 

Attempts have been made to model \textsc{tom} \cite{Yuan2021-jf,Rabinowitz2018-oc,Zhu2021-wq,Bara2021-ej} and some argue that Language Models have a degree of \textsc{tom} \cite{Ma2023-qc}. It's clear that \textsc{tom} is part of human cognition, and that it helps humans understand each other--a necessary part of learning and generating language that has meaning.

\section{Cognition, Emotion, and Embodiment}

Before infants learn to understand or speak their first words, they are able to communicate in another way: emotion. They cry when they need something or are uncomfortable, they smile when they are happy, and show other emotional states that signal to others what they are feeling. Only later do children start making vocal sounds that are intended to express some kind of communicative intent, and later still are they able to speak their first words that carry some kind of meaningful content. 

The question then is: as a person becomes more competent in language use, which is a cognitive process, do they move beyond emotion? If we are to believe what some philosophers have said about language being logical (leading to formal semantics), it seems to be the case that many believe that language and emotion are separate and distinct. Indeed, the more we can separate thinking from emotion, the better. This belief went so far as to depict an android (i.e., a humanoid robot) on \textit{Star Trek: The Next Generation} as a highly capable and intelligent being, but completely without emotion (that came later). Is that how we should view what it means to think and use language?

At this point, I believe, the answer is no. The meaning of many words includes emotional connotation. Pick any word, for example \emph{democracy} or \emph{career} and humans minimally attach positive or negative valence as part of the word meanings. This is the case particularly (perhaps surprisingly) for abstract words \cite{Lane2002-pk,Ponari2018-ve,Vigliocco2014-km}.\footnote{Related to emotion, \textit{affect} describes basic feelings of unpleasant to pleasant (valence) and from agitated to calm (arousal), and, like vision, is something into which language models could potentially be grounded, whereas emotion is more nuanced and tied to the abstract linguistic system \cite{Barrett2017-uh}. For clarity and consistency, we opt for the term emotion over affect.} As explained by \citet{Locke1995-tz}, without emotion there would be no interest, no need, no motivation and, consequently, questions or problems would never be posed, and there would be no intelligence. 

Locke (citing \citet{Alan_Sroufe2009-wa}) states that cognitive advances ``promote exploration, social development, and the differentiation of affect; and affective-social growth leads cognitive development [...] neither the cognitive nor the affective system can be considered dominant or more basic than the other; they are inseperable manifestations of the same integrated process [...] It is as valid to say that cognition is in the service of affect as to say that affect reflects cognitive processes."

Separating emotion from language may make language easier to model computationally, but the model will only have an approximation of linguistic meaning. Some have attempted to model emotional content inferred from text \cite{Alhuzali2018-hu,Murthy2021-ke,Saravia2018-zu,Xu2018-jl}, but the embodied, emotional content that is part of the meanings of words is not captured in the text itself. In other words, part of addressing the \emph{Symbol Grounding Problem} means grounding language into emotions \cite{Harnad1990-fr,Moro2020-kx}.

\citet{Smith2005-qt} showed that babies' experience of the world is profoundly multimodal. Every human who has ever lived and learned language has had a body in which to house the brain where language is processed. The brain functions and controls the agent that acts in a shared environment with everyone else. That agent is a human body. If we include things like sensory inputs where language needs to ground (indeed, must first ground), and the fact that emotion and cognition are intertwined, then is it the case that embodiment is required for language learning? Many think so \cite{Lakoff1999-cb,Johnson2008-zd,Di_Paolo2018-ba} including me \cite{Kennington2023-ft}. 

Without perception how could we learn that words refer to things? Without a body how could we enact actions that verbs denote, such as kick, walk, or swallow? Without a body how could we feel emotions that are tied to the connotation of many words? Without a body, how could we interact directly with others to learn sound patterns and our first words? According to scientists who adhere to embodied cognition, the answer to all of these is the same: we could not.




\section{Conclusion: Understanding Meaning}

The meaning of a word can be described using definitions, but the \textit{meaningfulness of language lies in the fact that it is about the world } \cite{Dahlgren1976-oh} and to be meaningful, something needs to be meaningful to something else. Words like \textit{hungry} and \textit{chair} are meaningful to me because I have experienced them in different ways. Hearing others say words like \textit{want} or \textit{please} means I have a degree of theory of mind, and I feel emotional valence when I hear or use certain words like \textit{tired} and \textit{coffee}. I have learned about words and what they mean to me because I have curiously explored the world over the years, and I could not have done that without a body that can perceive and act in the world. 

Does all of this mean that language models are not proper users of language? Well, our definition of language is \textit{symbolic communication}, and language models definitely do that. In fact, language models \textit{only} do that. The rest of the definition included repetition, change, and update of use--language models have been able to do that since before 2022. Language models are able to process language, though it we've learned anything about semantics and meaning, it is that those words that language models process probably aren't meaningful to them. Has a language model curiously explored the world? Has a language model felt the relief of drinking water to quench a thirst or sat in a chair after spending hours walking? Language models can pick up through text how to talk about things like quenching thirst and relief that sitting can bring, but they've never experienced those things themselves. 

My concluding remark is that language models can `understand' language abstractly, but lack of embodied, emotional experience means they are at a disadvantage when it comes to understanding the deeper meanings of the words that they process. In their recent paper, \citet{Beuls2024-xv} make an empirical case that Language Models that are trained only on text are missing important linguistic knowledge that could only be acquired in physical, person-to-person spoken interaction, arguing for, as I do here, a model of computational semantics that follows a similar curriculum to that of humans.  

What about the models themselves? Are they `properly' learning the semantics of language? \citet{Marcus2020-mg} argues that the path forward requires a ``hybrid, knowledge-driven, reasoning-based approach, centered around cognitive models." Thus a push for neuro-symbolic AI is underway, a version of AI where the power of formal systems is integrated into LMs trained on data with an increasing number of papers claiming they are addressing the neuro-symbolic challenge. We are thus still left with two semantic problems: the symbol grounding problem, and the neuro-symbolic problem, and work still needs to be done on solving both of those problems. A potential path forward could be that both are solved simultaneously using symbolic representations that ground into aspects of the physical world, yet are part of transformer LMs in the distributional representations (i.e., the embedding layer---connotation) as well as combined as done in other multimodal/visual LMs in the attention layers (i.e., denotation; see \citet{Kennington2025-cq}). In other words, simultaneously unifying distributional, grounded, and formal semantics is not only the best path forward technically, it is also the most likely theoretical answer. 

\section*{Acknowledgments}
Thanks to Annemarie Friedrich for her very helpful feedback. 

\bibliography{paperpile}
\bibliographystyle{acl_natbib}




\end{document}